\documentclass[sigconf,anonymous=False,review=False,natbib=True]{acmart}
\renewcommand\footnotetextcopyrightpermission[1]{} 
\usepackage{algorithm}
\usepackage{algorithmic}
\usepackage{tabularx, booktabs}
\usepackage{graphicx}
\usepackage{multirow}
\usepackage{ulem}
\usepackage{setspace}
\usepackage{xcolor}
\usepackage{array}
\usepackage{balance}
\usepackage{enumitem}
\usepackage{booktabs}
\usepackage{ulem}
\usepackage{CJKutf8}
\usepackage{makecell}
\usepackage{geometry}
\usepackage{subcaption}
\usepackage{hyperref}
\usepackage{stfloats}

\usepackage{caption}
\usepackage{graphicx}
\usepackage{float} 
\usepackage{subcaption}

\usepackage{listings}
\usepackage{xcolor}

\definecolor{codegreen}{rgb}{0,0.6,0}
\definecolor{codegray}{rgb}{0.5,0.5,0.5}
\definecolor{codepurple}{rgb}{0.58,0,0.82}
\definecolor{backcolour}{rgb}{0.95,0.95,0.92}

\lstdefinestyle{mystyle}{
    commentstyle=\color{codegreen},
    keywordstyle=\color{magenta},
    numbers=none,
    stringstyle=\color{codepurple},
    basicstyle=\ttfamily\footnotesize,
    breakatwhitespace=false,         
    breaklines=true,                 
    captionpos=b,                    
    keepspaces=true,                 
    numbersep=5pt,                  
    showspaces=false,                
    showstringspaces=false,
    showtabs=false,                  
    tabsize=2,
    breaklines=true,
    frame =lines
}

\usepackage{acronym}

\acrodef{IR}{Information Retrieval}
\acrodef{MLP}{multilayer perceptron}
\acrodef{SERP}{Search Engine Result Page}
\acrodef{ERP}{event related potential}
\acrodef{EEG}{electroencephalogram}
\acrodef{DT}{Gradient Boosting Decision Tree}
\acrodef{SST}{SST-EmotionNet}
\acrodef{AUC}{Area Under Curve}
\acrodef{IN}{Information Need}
\acrodef{fMRI}{functional magnetic resonance imaging}
\acrodef{BCI}{brain–computer interface}
\acrodef{BP}{band power}
\acrodef{DE}{differential entropy}
\acrodef{RASM}{rational asymmetry}
\acrodef{DASM}{differential asymmetry}
\acrodef{SVM}{support vector machines}
\acrodef{DBN}{deep belief networks}
\acrodef{KNN}{k-Nearest Neighbors}
\acrodef{HBG}{Height-Biased Gain}
\acrodef{GNN}{Graph Neural Network}
\acrodef{CNN}{Convolutional Neural Network}

\AtBeginDocument{%
  \providecommand\BibTeX{{%
    \normalfont B\kern-0.5em{\scshape i\kern-0.25em b}\kern-0.8em\TeX}}}

\makeatletter
\def\hlinew#1{%
  \noalign{\ifnum0=`}\fi\hrule \@height #1 \futurelet
   \reserved@a\@xhline}

\def\@ACM@checkaffil{
    \if@ACM@instpresent\else
    \ClassWarningNoLine{\@classname}{No institution present for an affiliation}%
    \fi
    \if@ACM@citypresent\else
    \ClassWarningNoLine{\@classname}{No city present for an affiliation}%
    \fi
    \if@ACM@countrypresent\else
        \ClassWarningNoLine{\@classname}{No country present for an affiliation}%
    \fi
}
\makeatother

\setcopyright{acmcopyright}
\copyrightyear{2021}
\acmYear{2021}
\acmDOI{10.1145/1122445.1122456}

\acmConference[]{
}{}{}

\begin{document}

\title{GNN4EEG: A Benchmark and Toolkit for Electroencephalography Classification with Graph Neural Network}


\author{Kaiyuan Zhang}
\email{kaiyuanzhang2001@gmail.com}
\affiliation{SCIT, Beijing Jiaotong University}

\author{Ziyi Ye}
\email{yeziyi1998@gmail.com}
\affiliation{BNRist, DCST, Tsinghua University}

\author{Qingyao Ai}
\email{aiqy@tsinghua.edu.cn}
\affiliation{BNRist, DCST, Tsinghua University}

\author{Xiaohui Xie}
\email{xiexh_thu@163.com}
\affiliation{%
  \institution{BNRist,DCST,Tsinghua University}
  \city{Beijing}
  \country{China}}

\author{Yiqun Liu}
\email{yiqunliu@tsinghua.edu.cn}
\affiliation{%
  \institution{BNRist,DCST,Tsinghua University}
  \city{Beijing}
  \country{China}}

\renewcommand{\shortauthors}{Zhang, et al.}


\begin{abstract}

Electroencephalography~(EEG) classification is a crucial task in neuroscience, neural engineering, and several commercial applications. 
Traditional EEG classification models, however, have often overlooked or inadequately leveraged the brain's topological information. 
Recognizing this shortfall, there has been a burgeoning interest in recent years in harnessing the potential of Graph Neural Networks ~(GNN) to exploit the topological information by modeling features selected from each EEG channel in a graph structure. 
To further facilitate research in this direction, we introduce GNN4EEG, a versatile and user-friendly toolkit for GNN-based modeling of EEG signals.
GNN4EEG comprises three components: (i)~A large benchmark constructed with four EEG classification tasks based on EEG data collected from 123 participants. (ii)~Easy-to-use implementations on various state-of-the-art GNN-based EEG classification models, e.g., DGCNN, RGNN, etc. (iii)~Implementations of comprehensive experimental settings and evaluation protocols, e.g., data splitting protocols, and cross-validation protocols.
GNN4EEG is publicly released at \href{https://github.com/Miracle-2001/GNN4EEG}{https://github.com/Miracle-2001/GNN4EEG}.

\end{abstract}

\settopmatter{printacmref=false}

\ccsdesc[500]{Computing methodologies~Machine learning}
\ccsdesc[500]{Information systems~Multimedia information systems}

\keywords{Experimentation,
Graph Neural Network,
Electroencephalography}


\maketitle
\vspace{-0.5mm}
\section{introduction}
Electroencephalogram (EEG) classification has become an emerging and increasingly remarkable task in neuroscience~\cite{srinivasan2007cognitive}, neural engineering~\cite{hsu2010eeg,jia2020graphsleepnet}, and even several existing commercial applications~\cite{vecchiato2011spectral,baumgartner2006neural}.
Through multiple channels placed on the scalp, EEG devices can measure the electrical changes in voltage that occur in the brain.
The challenge of EEG classification lies in how to handle this multi-channel time-series data of electrical changes.
Traditionally, EEG classification models transform the multi-channel time-series data into a one-dimensional or two-dimensional format and then put one-dimensional data into classifiers such as \ac{SVM} or \ac{MLP}~\cite{chatterjee2016eeg,lin2007multilayer}, and two-dimensional data into \ac{CNN}~\cite{dai2019eeg,craik2019deep}.
The reduction of data dimensionality may lead to the loss of inter-channel interactions in EEG data, resulting in a decline in classification performance.

Recently, there has been a notable surge in the employment of Graph Neural Networks~(GNNs) for the classification of EEG data. This paradigm is rapidly gaining favor within the research community~\cite{jia2020graphsleepnet, song2018eeg, zhong2020eeg}.
Different from \ac{MLP} or \ac{CNN} based methods, \ac{GNN} based methods treat each channel as a node in the computing graph and can learn their interactions flexibly, as shown in Figure~\ref{GNN}.
To address unique challenges in EEG analysis, especially the spatial structural features extracted from EEG, researchers have proposed several variants of GNN models tailored for EEG classification, such as DGCNN~\cite{song2018eeg}, RGNN~\cite{zhong2020eeg}, SparseDGCNN~\cite{zhang2021sparsedgcnn}, etc.
However, it remains challenging to evaluate the transferability of these models and implement GNN-based EEG classification models in practice due to the lack of easy-to-use toolkits and large-scale public benchmarks.
To tackle this, we build GNN4EEG, a benchmark and toolkit for EEG classification with \ac{GNN}.
GNN4EEG is composed of three components:

\begin{figure}[]
\centering 
\includegraphics[height=5cm,width=0.85\columnwidth]{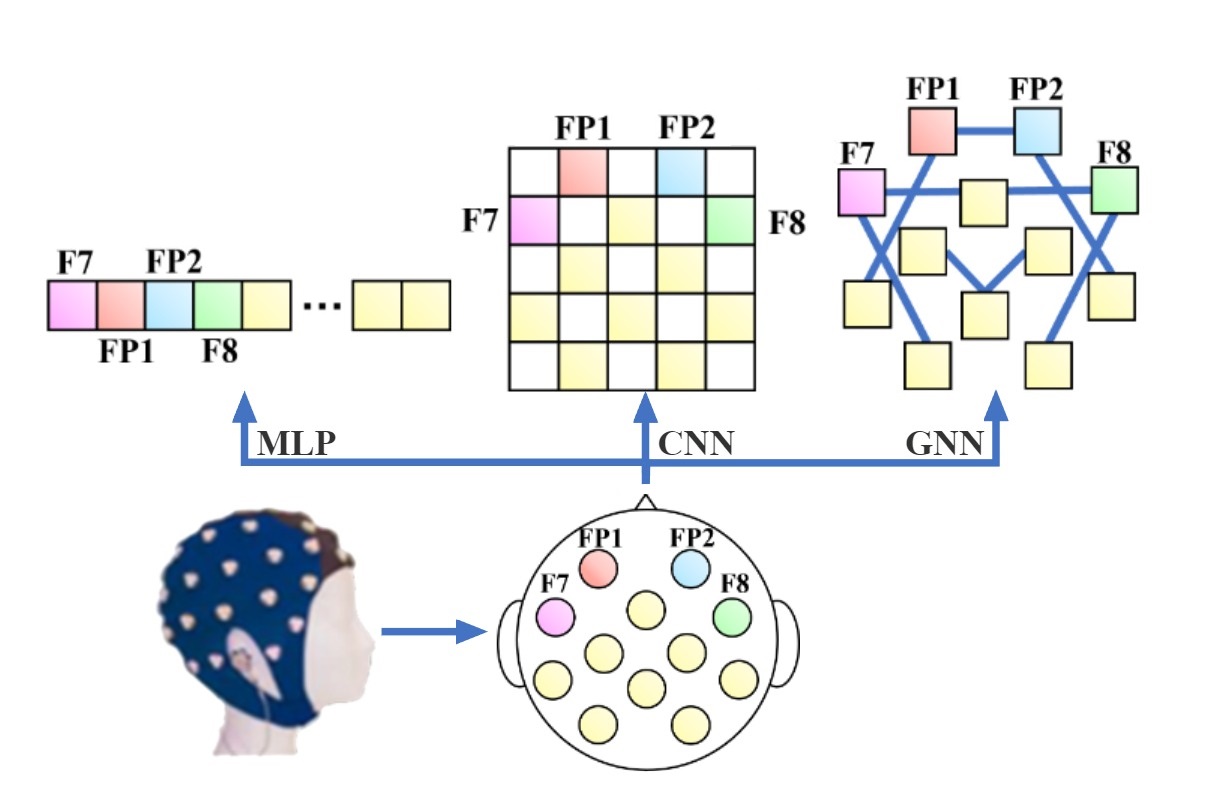}
\caption{Three typical models for EEG classification. }
\label{GNN}
\end{figure}


\textbf{Benchmark.}
We built a large-scale benchmark with the Finer-grained Affective Computing EEG Dataset~(FACED)~\cite{FACED}. As far as we know, FACED is the largest affective computing dataset, which is constructed by recording 32-channel EEG signals from a large cohort of 123 subjects watching 28 emotion-elicitation video clips, different from existing EEG benchmarks~\cite{zhang2021eegdenoisenet,duan2013differential} which are only collected from a relatively small size, e.g., 40 participants in DEAP~\cite{koelstra2011deap}, 15 participants in SEED and SEED-IV~\cite{duan2013differential}.
After standardized and unified data preprocessing, we devised four EEG classification tasks based on FACED as our benchmark to explore the effectiveness of EEG classification models in different evaluation protocols.

\textbf{Codebase for GNN-based EEG classification.}
We develop an open-source software toolkit to support the adaptation and evaluation of frequently used GNN-based EEG classification models, including DGCNN~\cite{song2018eeg}, RGNN~\cite{zhong2020eeg}, SparseDGCNN~\cite{zhang2021sparsedgcnn}, and HetEmotionNet~\cite{jia2021hetemotionnet}.
To increase the scalability of our toolkit,  we also provide easy-to-use interfaces to implement customized EEG classification models or change backbone GNN algorithms in existing EEG classification models.

\textbf{Robust Evaluation}
To compare different models under different settings, GNN4EEG implements different evaluation protocols, including the data splitting protocols and validation protocols.
The data splitting protocols include intra-subject and cross-subject splitting protocols~\cite{FACED} to explore the classification performance without and with the issue of subject specificity, respectively.
In previous studies, cross-validation~\cite{golub1979generalized} is widely used as the validation protocols~\cite{song2018eeg, zhong2020eeg, jia2021hetemotionnet}.
However, the implementation of existing cross-validation~\cite {jia2021hetemotionnet} typically permits varying selections of the number of training epochs for different training folds to achieve optimal validation performance.
This gives us the highest validation performance but is unreasonable since the validation performance varies with the number of training epochs. 
To tackle this issue, we implement three validation protocols, which use different or fixed selections on the number of training epochs among each validation fold or the number of training epochs selected by a nested cross-validation process~\cite{krstajic2014cross}.
Moreover, the proposed toolkit also extends the automatic hyper-parameter tuning during nested cross-validation for real-life adaptation scenarios, which can proceed the whole process of training and tuning an available EEG classification model for real-time applications in just a few lines of code.




\section{Toolkit Framework}
GNN4EEG is an integrated benchmark and toolkit which implements 4 EEG classification tasks as the benchmark, 3 validation protocols, and 4 GNN models. 

\vspace{-0.5mm}
\subsection{Benchmark Construction}
Our benchmark construction is based on the FACED dataset~\cite{FACED}, the labels of which can be divided into 9 dimensions~(amusement, inspiration, joy, tenderness; anger, fear, disgust, sadness, and neutral emotion) or in 2 dimensions~(positive and negative).

\textbf{Data preprocessing.} We preprocess the dataset following the example provided in the original paper~\cite{FACED}. Specifically, we drop the last 2 channels and preserve the last 30 seconds epoch for each video clip. Afterward, the differential entropy~(DE) ~\cite{duan2013differential} features smoothed by linear dynamic systems~(LDS)~\cite{gajic2003linear} are applied to extract five frequency bands (delta, theta, alpha, beta, and gamma) features~\cite{sanei2013eeg} for each second of EEG signals~(without overlapping) in each channel.

\textbf{Data Splitting.} Whether the data collected from an individual appearing in the validation set also appear in the training set yields two different data splitting methods, i.e., intra-subject and cross-subject.

In the intra-subject task, each subject's EEG signals appear both in the training and validation sets. On the contrary, in cross-subject tasks, the training set contains EEG signals from only part of the subjects, while other subjects' EEG signals are divided into validation sets. 
We have created four benchmarks using the dataset, namely cross-9, cross-2, intra-9, and intra-2. 
These benchmarks were constructed based on intra-subject and cross-subject data splitting settings and the emotional labels of 9 and 2 dimensions, respectively.

\subsection{GNN for EEG Classification}
In this section, we give a brief introduction to graph convolution network~(GCN)~\cite{wu2019simplifying}, a typical type of GNN that uses convolutional layers to aggregate information~\cite{xu2018powerful}, as well as the four GNN-based EEG classification models implemented in our toolkit, i.e., DGCNN, RGNN, SparseDGCNN, and HetEmotionNet.

Formally, a graph can be defined as $G=(V, E)$, where $V$ denotes the node set and $E$ denotes the edge set. 
In EEG classification tasks, the node set refers to the multiple channels, and the edge set denotes their topographical relations.
The input feature can be represented as a matrix $X\in \mathbb{R}^{n \times d}$, where $n=|V|$ and $d$ denote the dimension of input features. 
The edge weight of $E$ can be represented by a weighted adjacency matrix $A\in \mathbb{R}^{n\times n}$ with self-loops. 
Furthermore, we can calculate the diagonal degree matrix $D$ of $A$ via $D_{ii}=\sum_j A_{ij}$. 

\subsubsection{GCN}
GCN is useful when aggregating and combining features extracted from EEG signals end-to-end. In general, the forward propagation of one GCN layer can be written as follows~\cite{wu2019simplifying}.
\begin{equation}
\label{GCN_forward}
H^{l+1}=\sigma(D^{-\frac{1}{2}}AD^{-\frac{1}{2}}H^lW^l) 
\end{equation}
where $l=0,1,...,L-1,L$ denotes the number of layers, with $H^0=X$. For each layer $l$, $\sigma$ represents a non-linear function, and $W^l$ represents a weighted matrix.

\subsubsection{DGCNN}
DGCNN~\cite{song2018eeg} model consists of a graph filtering layer~(can be approximated as Eq.\ref{GCN_forward}), a convolution layer, an activation layer and a fully connected layer. The adjacency matrix $A$ in DGCNN is optimized during training, and $A_{ii}\ge 0$ is always held. Though DGCNN is simple enough, it usually outperforms other models like SVM~\cite{boser1992training} and GSCCA~\cite{zhang2020fusing} because it exploits the topographical information with GNN.

\subsubsection{RGNN}
To deal with cross-subject issues and noisy labels in EEG emotion recognition tasks, RGNN~\cite{zhong2020eeg} proposes two regularizers named NodeDAT and EmotionDL. Unlike DGCNN, RGNN introduced a negative number to represent the global connection. 

\subsubsection{SparseDGCNN}
Different from RGNN, SparseDGCNN~\cite{zhang2021sparsedgcnn} aims to improve the sparseness of the adjacency matrix during training by imposing a sparseness constraint. Moreover, this model also applies a forward-backward splitting method to guarantee the convergence of the training process. 

\subsubsection{HetEmotionNet}
HetEmotionNet~\cite{jia2021hetemotionnet} uses a novel two-stream heterogeneous graph recurrent neural network to capture both spatial-spectral and spatial-temporal features in multi-modal physiological signals. Graph Transformer Network~(GTN)~\cite{yun2019graph} and Gated Recurrent Unit~(GRU)~\cite{chung2014empirical} are adopted for heterogeneity graph aggregation and recurrent features extraction, respectively.

\subsection{Validation Protocols}
Our toolkit implements 3 validation protocols, i.e., cross-validation, fixed epoch cross-validation and nested cross-validation. It is worth pointing out that although the descriptions and pseudo-code below only show the tuning method of epoch numbers, other parameters like learning rate and hidden layer dimension can also be automatically tuned in our toolkit. 

\subsubsection{Cross Validation (CV)} Cross validation~\cite{golub1979generalized} is a typical re-sampling method. Conventionally, the data is divided into $K$ parts, and the model is evaluated in $K$ training and validating iterations. On the $i^{th}$ iteration, the $i^{th}$ part of the data is chosen as the validation set and others as the training set. As implemented in prior work~\cite{jia2021hetemotionnet}, the maximum validation accuracy among all epochs is regarded as the evaluation result of the $i^{th}$ iteration. 

Formally, assume $Acc_{i,j}$ as the validation accuracy of the $j^{th}$ epoch in the $i^{th}$ iteration. Then, for each iteration $i$, compute $EPO_i=argmax\{{Acc_{i,1},Acc_{i,2},...,Acc_{i,T}}\}$ where $T$ is the number of epoch and $MAX_i=Acc_{i,EPO_i}$. Finally, CV reports the averaged maximum accuracy among all iterations, i.e., $\frac{1}{K}\sum_{i=1}^{K}MAX_i$.



\subsubsection{Fixed Epoch Cross Validation (FCV)} 
Note that in the cross-validation method, $EPO_*$ can vary for different iterations. 
However, due to the performance fluctuation in different epochs, the accuracy acquired by simply choosing the best training epoch for each fold may introduce unpredictable random factors.

To address the above issue, a fixed epoch cross-validation is proposed, in which a fixed number of epochs is selected among all iterations, i.e., $EPO_*$ is the same for all folds. Specifically, after acquiring $Acc_{i,j}$, FCV computes $AVG_j=\frac{1}{K}\sum_{i=1}^{K}Acc_{i,j}$ for each epoch. Afterwards, $max\{{AVG_1,AVG_2,...,AVG_T}\}$ is reported as the final result.

\subsubsection{Nested Cross Validation (NCV)} Evaluation methods above choose the hyper-parameters~(epoch number) according to the performance on the validation set, but still report the accuracy on the same data set, thus leading to data leakage. So in order to evaluate the generalization ability more appropriately, nested cross-validation is applied~\cite{krstajic2014cross}.

Based on CV, NCV further divides the training set of each iteration into another $K'$ equal parts, introducing an "inner" and "outer" cross-validation manner. The "inner" selects the best epoch number, and the "outer" averages the accuracy over $K$ folds. Namely, NCV introduces not only a training set and a validation set but also a test set, thus providing a more reliable estimate of the model's generalization ability. The pseudo-code is shown in Algorithm~\ref{"alg.NCV"}.

\begin{algorithm}
	\renewcommand{\algorithmicrequire}{\textbf{Input:}}
	\renewcommand{\algorithmicensure}{\textbf{Output:}}
	\caption{The procedure of NCV}
        \label{"alg.NCV"}
	\begin{algorithmic}[1]
            \REQUIRE{Dataset $\mathcal{D}$, GNN model $\mathcal{M}$, number of equal parts $K$, number of inner partition $K'$, number of epochs $T$, other hyper-parameters $P$}
            \ENSURE{The averaged test accuracy among all $K$ iterations.}
            \FOR{$i=1$:$K$}
            \STATE Split $\mathcal{D}$ into 
            $\mathcal{D}^{test}_i$, $\mathcal{D}^{train{\&}val}_i$ 

            \FOR{$j=1$:$K'$}
            \STATE Initialize $\mathcal{M}$ via hyper-parameters $P$
            \STATE Split $\mathcal{D}^{train{\&}val}_i$  into 
            $\mathcal{D}^{train}_j$, $\mathcal{D}^{val}_j$ 
            \FOR{$t=1$:$T$}
             \STATE Train $\mathcal{M}$ on $\mathcal{D}^{train}_j$ for an epoch
             \STATE Compute the validation accuracy $Acc^{val}_{j,t}$ for $\mathcal{M}$ on  $\mathcal{D}^{val}_j$ 
            \ENDFOR
            \ENDFOR
            \STATE Compute $AVG_t=\frac{1}{K'}\sum_{j=1}^{K'}Acc^{val}_{j,t}$ for $t=1,2,...,T$
            \STATE Compute $T'=argmax\{{AVG_1,AVG_2,...,AVG_T}\}$
            
            \STATE Initialize $\mathcal{M}$ via hyper-parameters $P$
            \STATE Train $\mathcal{M}$ on $\mathcal{D}^{train{\&}val}_i$  for $T'$ epochs.
            \STATE Compute the test accuracy $Acc^{test}_i$ for $\mathcal{M}$ on $\mathcal{D}^{test}_i$ 
            \ENDFOR
		\STATE Compute $AVG=\frac{1}{K}\sum_{i=1}^{K}Acc^{test}_i$
		\RETURN  $AVG$
	\end{algorithmic}  
\end{algorithm}

\vspace{-0.5mm}
\subsection{Example Usage}
In this section, we describe the usage of our toolkit, including data splitting, model selection, and validation protocol configuration. Note that data preprocessing methods are not provided, so users should apply these methods on their own if necessary. 

\subsubsection{Data Splitting}
First, it is necessary to choose the data splitting protocols, i.e., intra-subject or cross-subject. A list describing the subject of each sample should be provided to guide the splitting. 

\vspace{-1mm}

\subsubsection{Model Selection}
To initiate a specific model, parameters like the number of classification categories, graph nodes, hidden layer dimension, and GNN layers should be included. Electrode positions, frequency values, and other options are also necessities for certain GNN models. 


\subsubsection{Validation Protocols and Other Training Configurations}
The final step is to declare the validation protocols and other configurations. As illustrated above, GNN4EEG provides three validation protocols, i.e., CV, FCV, and NCV. For detailed training configurations, the user can set the learning rate, dropout rate, number of epochs, $L_1$ and $L_2$ regularization coefficient, batch size, optimizer, and training device.  


To better illustrate the toolkit, an example of cross-9 NCV evaluation via the DGCNN model is shown as follows:

\begin{lstlisting}[language=Python]
# Define the data splitting protocol
loader=ge.protocols.data_split("cross_subject",data_samples,data_labels,subject_id_list)
# Select the DGCNN model and some fixed hyper-parameters
model=ge.models.DGCNN(num_of_nodes,num_of_layers,electrode_position)
# Declare the validation protocol and the tuning range to start the evaluation 
result=ge.protocols.evaluation("ncv",K=10,K_inner=3,grid={"hiddens":[20,40],"lr":[1e-4,1e-3],"epoch":list(range(1,101))},categories=9,model,loader)
\end{lstlisting}


\vspace{-1mm}
\section{experiment}
In this section, we present the experimental setup and the evaluation results using the proposed GNN4EEG toolkit. Analyses of overall performances and sensitiveness are also elaborated. The experiments are implemented on NVIDIA GeForce RTX 3090, and code has been released~\footnote{\href{https://github.com/Miracle-2001/GNN4EEG}{https://github.com/Miracle-2001/GNN4EEG.}}. 


\vspace{-7mm}
\subsection{Experimental Setup}
We set the fold number $K=10$ for all validation protocols and the ``inner'' fold number $K'=3$ for NCV. In intra-subject tasks, the 30 seconds EEG signals among all video clips and subjects are equally split into $K$ folds. While in cross-subject tasks, the 123 subjects are split into $K$ folds, with the last fold containing 15 subjects and the former each containing 12 subjects. 

We tune the number of hidden dimensions from $\{20,40,80\}$ and the learning rate from $\{0.0001,0.001,0.01\}$ for all tasks and models. Moreover, the dropout rate is $0.5$, the number of GNN layers is $2$, the batch size is $256$, and the maximum number of epochs is set as $100$. 
To address potential overfitting in different settings, we have utilized different weights for the $L_1$ and $L_2$ norm in different tasks. 
Specifically, both weights are set as $0.001$ for intra-2, $0.005$ for cross-9, and $0.003$ for cross-2 and intra-9.

\begin{table*}[]
\caption{Accuracy under different configurations. The best-performing models are highlighted with boldface.}
\label{tab_overal}
\centering
\tabcolsep=0.016\linewidth
\begin{tabular}{|c|ccc|ccc|ccc|ccc|}
\hline
\multicolumn{1}{|c|}{\multirow{2}{*}{Model}} & \multicolumn{3}{c|}{intra-2} & \multicolumn{3}{c|}{cross-2} & \multicolumn{3}{c|}{intra-9} & \multicolumn{3}{c|}{cross-9} \\
\cline{2-13}

\multicolumn{1}{|c|}{} & CV & FCV & NCV & CV & FCV & NCV & CV & FCV & NCV & CV & FCV & NCV\\
\hline
DGCNN  & $72.9$ & 72.4 & 72.6 & $69.4$ & 66.9 & 65.1 & 32.8  & 33.5 & $33.6$ & $30.8$ & 28.9  & 26.7\\
RGNN  & $81.7$ & 81.1 & 81.4 & \textbf{72.2} & \textbf{70.2} & \textbf{68.5} & $40.6$ & 39.5 & 40.2  & \textbf{34.0} & \textbf{32.1} & \textbf{31.3} \\
SparseDGCNN & $72.8$ & 72.3 & 72.5 & $69.8$ & 66.6  & 65.1 & $33.8$  & 32.8 & 33.4 & $30.7$  & 29.0 & 27.7\\
HetEmotionNet & \textbf{94.5}& \textbf{92.8} & \textbf{93.3} & $69.2$ & 67.3 & 63.0 & \textbf{49.5}  & \textbf{48.6} & \textbf{44.9} & $32.5$ & 30.8 & 30.8\\

\hline
\end{tabular}
\end{table*}

\begin{figure}[]
	\centering
	\begin{subfigure}{0.48\columnwidth}
		\centering
		\includegraphics[width=1\linewidth]{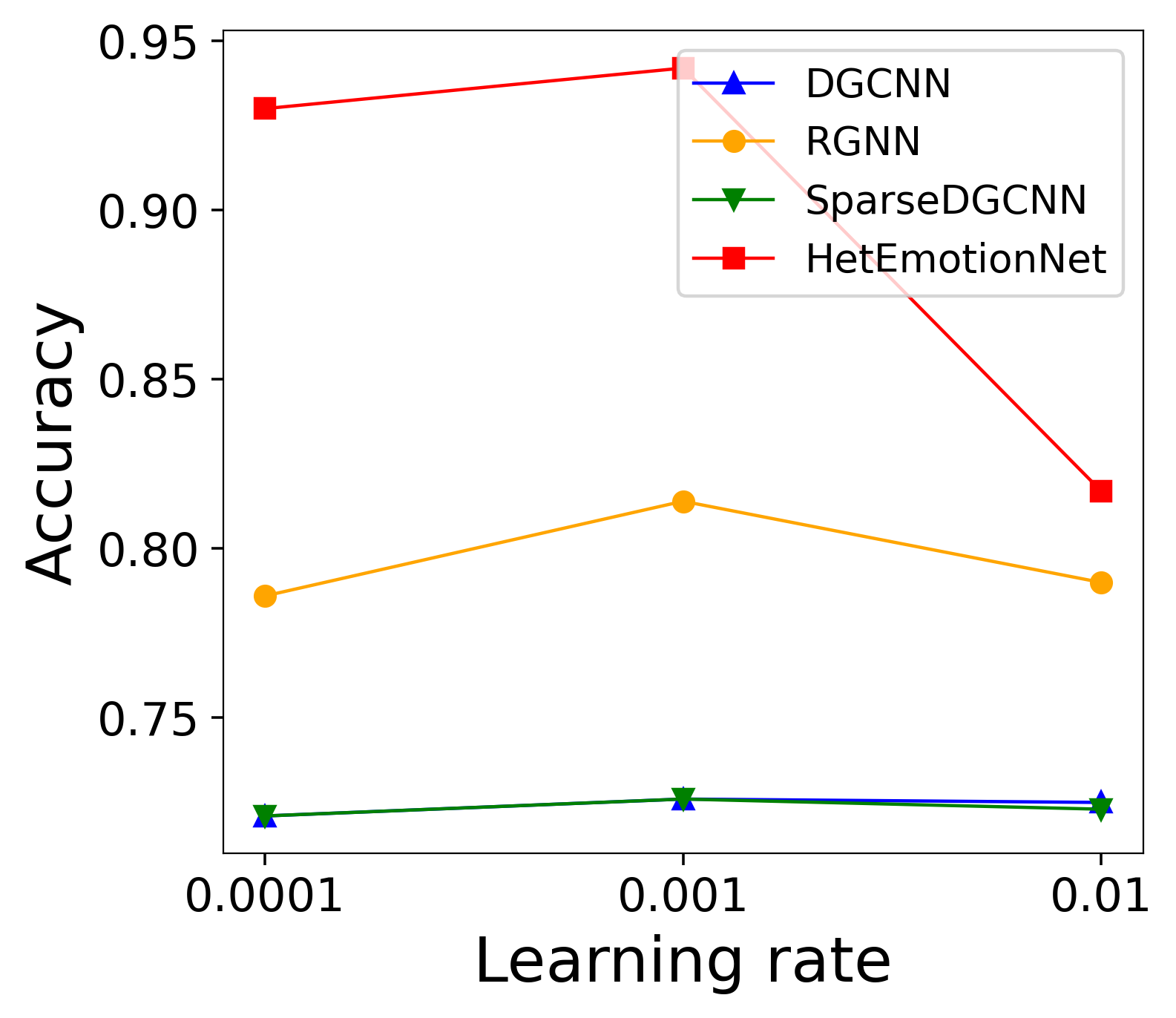}
		\caption{Intra-2}
		\label{intra-2-lr}
	\end{subfigure}
	\centering
	\begin{subfigure}{0.48\columnwidth}
		\centering
		\includegraphics[width=1\linewidth]{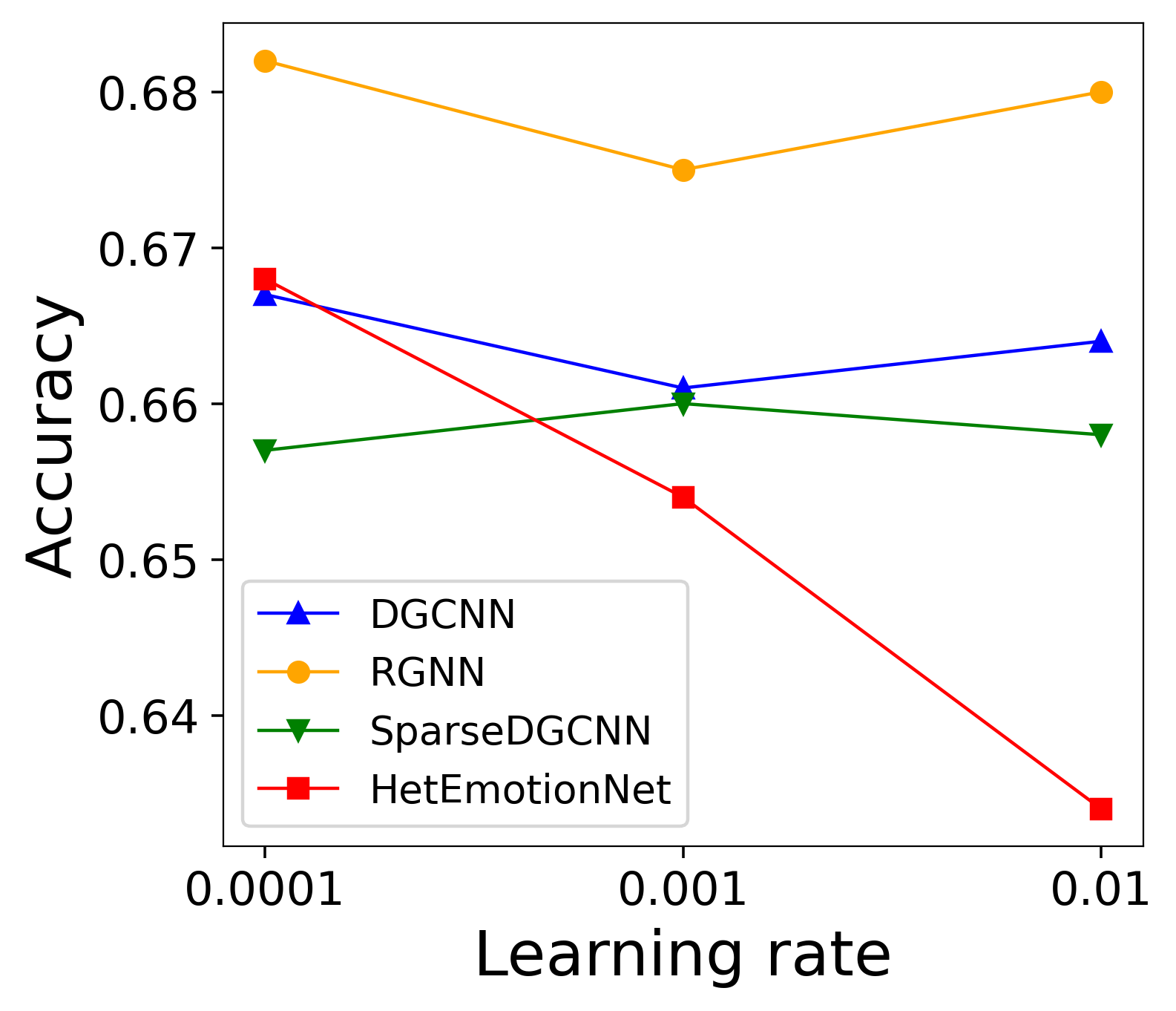}
		\caption{Cross-2}
		\label{cross-2-lr}
	\end{subfigure}
	\centering
	\begin{subfigure}{0.48\columnwidth}
		\centering
		\includegraphics[width=1\linewidth]{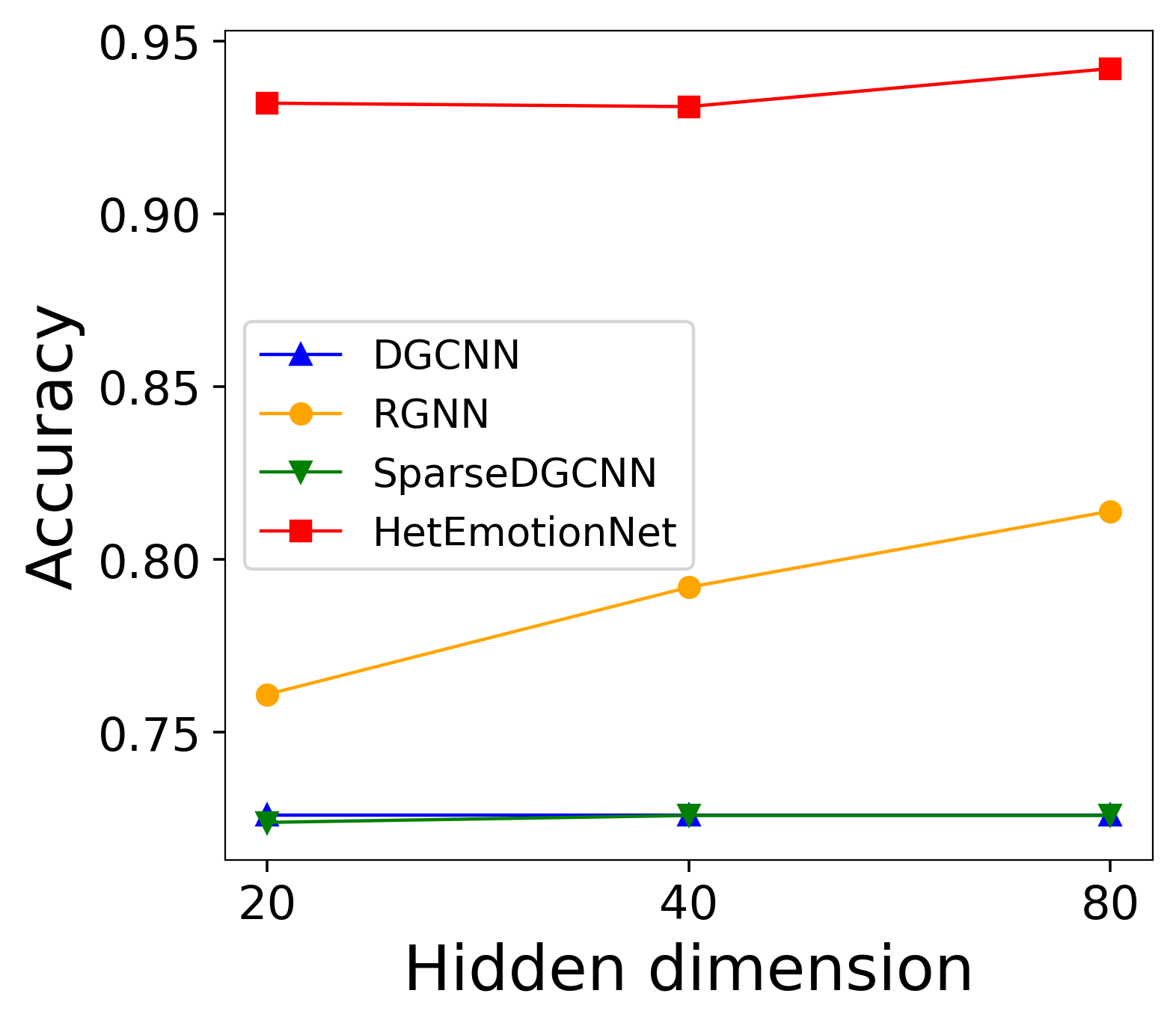}
		\caption{Intra-2}
		\label{intra-2-hid}
	\end{subfigure}
         \begin{subfigure}{0.48\columnwidth}
            \centering
            \includegraphics[width=1\linewidth]{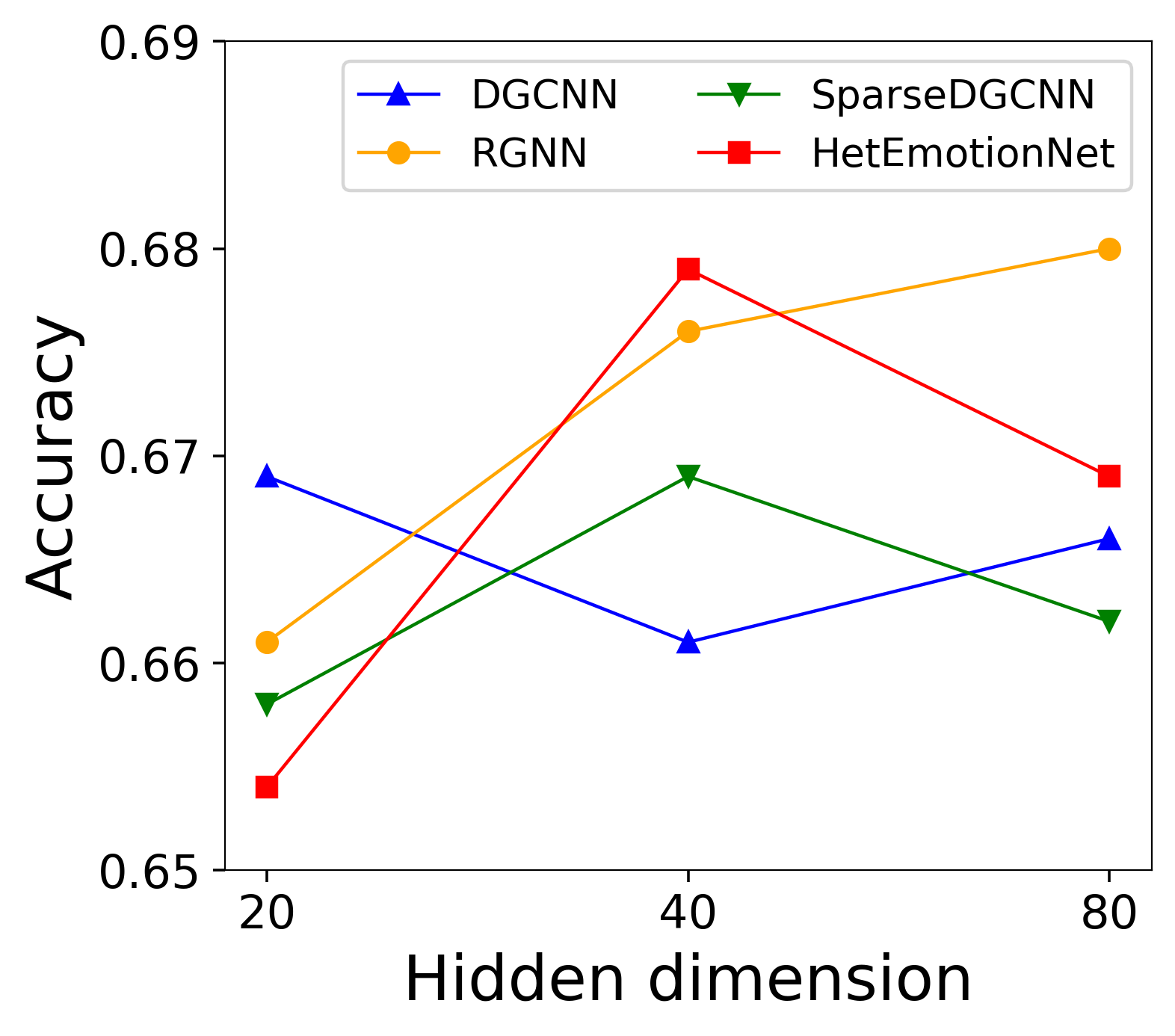}
            \caption{Cross-2}
            \label{cross-2-hid}
        \end{subfigure}
	\caption{Sensitivity analysis. When conducting sensitivity analysis on the hidden layer dimension or the learning rate, the other one is fixed at the value that yields the optimal average test set accuracy.}
	\label{sensitive}
\end{figure}

\subsection{Results and Analyses}
Based on the toolkit, we evaluate the performance of GNN models under the three validation protocols and four classification tasks. 
\subsubsection{Overall Results}
Table~\ref{tab_overal} shows the overall accuracy of each task. By extracting features from both spatial-spectral and spatial-temporal domains, HetEmotionNet outperforms other models in intra-subject tasks, especially in the intra-2 task. However, RGNN achieves the best accuracy in cross-subject tasks due to the application of the NodeDAT regularizer. SparseDGCNN has similar performance compared to DGCNN under this experimental setting. 
For validation protocols, CV performs slightly better than FCV because of the variety of the chosen epoch numbers on different folds.
Besides, to address data leakage, NCV reports the performances on the test set, leading to a poorer result most of the time compared to CV, whose performances are reported on the validation set. Although NCV performs worse than FCV in certain tasks, the rationality in measuring generalization ability still makes it a practical protocol.

\subsubsection{Sensitivity Analyses}
We explore the influences of different hyper-parameters on the testing accuracy of NCV validation through sensitivity analyses. Empirically, we find both hidden dimension number and learning rate have a relatively significant impact on the accuracy. By respectively fixing one of the two hyper-parameters according to the average performance on the test set, we show the accuracy with varying values of the other parameter as illustrated in Fig~\ref{sensitive}.

Generally, hyper-parameter variation has a lower influence in intra-subject tasks than cross-subject tasks. 
This is expected because EEG signals have significant differences between subjects, producing a relatively large gap between the training sets' distribution and the validation sets' distribution. 
 In intra-subject tasks, a larger hidden dimension usually generates better accuracy, which comes from the promotion of expressive ability. However, the observations are different in cross-subject tasks for similar reasons above. 
Regarding the learning rate, a distinct decrease in accuracy appears in HetEmotionNet when the value reaches 0.01, while other models are relatively robust. 
One possible explanation is that HetEmotionNet contains more parameters, while a lower learning rate can better guarantee convergence. 




\section{conclusion}
This paper proposes an easy-to-use open-source toolkit named GNN4EEG. 
GNN4EEG consists of four EEG classification benchmarks, four GNN-based EEG classification models, and three validation protocols. 
Experiments on a large 123-participant dataset are reported. 
Overall analyses and sensitivity analyses have been demonstrated to verify the practicability of our toolkit.

\bibliographystyle{ACM-Reference-Format}
\normalem
\balance
\bibliography{references}


\begin{thebibliography}{27}


\ifx \showCODEN    \undefined \def \showCODEN     #1{\unskip}     \fi
\ifx \showDOI      \undefined \def \showDOI       #1{#1}\fi
\ifx \showISBNx    \undefined \def \showISBNx     #1{\unskip}     \fi
\ifx \showISBNxiii \undefined \def \showISBNxiii  #1{\unskip}     \fi
\ifx \showISSN     \undefined \def \showISSN      #1{\unskip}     \fi
\ifx \showLCCN     \undefined \def \showLCCN      #1{\unskip}     \fi
\ifx \shownote     \undefined \def \shownote      #1{#1}          \fi
\ifx \showarticletitle \undefined \def \showarticletitle #1{#1}   \fi
\ifx \showURL      \undefined \def \showURL       {\relax}        \fi
\providecommand\bibfield[2]{#2}
\providecommand\bibinfo[2]{#2}
\providecommand\natexlab[1]{#1}
\providecommand\showeprint[2][]{arXiv:#2}

\bibitem[\protect\citeauthoryear{Baumgartner, Valko, Esslen, and J{\"a}ncke}{Baumgartner et~al\mbox{.}}{2006}]%
        {baumgartner2006neural}
\bibfield{author}{\bibinfo{person}{Thomas Baumgartner}, \bibinfo{person}{Lilian Valko}, \bibinfo{person}{Michaela Esslen}, {and} \bibinfo{person}{Lutz J{\"a}ncke}.} \bibinfo{year}{2006}\natexlab{}.
\newblock \showarticletitle{Neural correlate of spatial presence in an arousing and noninteractive virtual reality: an EEG and psychophysiology study}.
\newblock \bibinfo{journal}{\emph{CyberPsychology \& Behavior}} \bibinfo{volume}{9}, \bibinfo{number}{1} (\bibinfo{year}{2006}), \bibinfo{pages}{30--45}.
\newblock


\bibitem[\protect\citeauthoryear{Boser, Guyon, and Vapnik}{Boser et~al\mbox{.}}{1992}]%
        {boser1992training}
\bibfield{author}{\bibinfo{person}{Bernhard~E Boser}, \bibinfo{person}{Isabelle~M Guyon}, {and} \bibinfo{person}{Vladimir~N Vapnik}.} \bibinfo{year}{1992}\natexlab{}.
\newblock \showarticletitle{A training algorithm for optimal margin classifiers}. In \bibinfo{booktitle}{\emph{Proceedings of the fifth annual workshop on Computational learning theory}}. \bibinfo{pages}{144--152}.
\newblock


\bibitem[\protect\citeauthoryear{Chatterjee and Bandyopadhyay}{Chatterjee and Bandyopadhyay}{2016}]%
        {chatterjee2016eeg}
\bibfield{author}{\bibinfo{person}{Rajdeep Chatterjee} {and} \bibinfo{person}{Tathagata Bandyopadhyay}.} \bibinfo{year}{2016}\natexlab{}.
\newblock \showarticletitle{EEG based motor imagery classification using SVM and MLP}. In \bibinfo{booktitle}{\emph{2016 2nd International Conference on Computational Intelligence and Networks (CINE)}}. IEEE, \bibinfo{pages}{84--89}.
\newblock


\bibitem[\protect\citeauthoryear{Chen, Wang, Huang, Hu, Shen, and Zhang}{Chen et~al\mbox{.}}{2023}]%
        {FACED}
\bibfield{author}{\bibinfo{person}{Jingjing Chen}, \bibinfo{person}{Xiaobin Wang}, \bibinfo{person}{Chen Huang}, \bibinfo{person}{Xin Hu}, \bibinfo{person}{Xinke Shen}, {and} \bibinfo{person}{Dan Zhang}.} \bibinfo{year}{2023}\natexlab{}.
\newblock \showarticletitle{FACED: a large Finer-grained Affective Computing EEG Dataset}.
\newblock  (\bibinfo{year}{2023}).
\newblock
\urldef\tempurl%
\url{https://doi.org/10.7303/syn50614194}
\showDOI{\tempurl}


\bibitem[\protect\citeauthoryear{Chung, Gulcehre, Cho, and Bengio}{Chung et~al\mbox{.}}{2014}]%
        {chung2014empirical}
\bibfield{author}{\bibinfo{person}{Junyoung Chung}, \bibinfo{person}{Caglar Gulcehre}, \bibinfo{person}{KyungHyun Cho}, {and} \bibinfo{person}{Yoshua Bengio}.} \bibinfo{year}{2014}\natexlab{}.
\newblock \showarticletitle{Empirical evaluation of gated recurrent neural networks on sequence modeling}.
\newblock \bibinfo{journal}{\emph{arXiv preprint arXiv:1412.3555}} (\bibinfo{year}{2014}).
\newblock


\bibitem[\protect\citeauthoryear{Craik, He, and Contreras-Vidal}{Craik et~al\mbox{.}}{2019}]%
        {craik2019deep}
\bibfield{author}{\bibinfo{person}{Alexander Craik}, \bibinfo{person}{Yongtian He}, {and} \bibinfo{person}{Jose~L Contreras-Vidal}.} \bibinfo{year}{2019}\natexlab{}.
\newblock \showarticletitle{Deep learning for electroencephalogram (EEG) classification tasks: a review}.
\newblock \bibinfo{journal}{\emph{Journal of neural engineering}} \bibinfo{volume}{16}, \bibinfo{number}{3} (\bibinfo{year}{2019}), \bibinfo{pages}{031001}.
\newblock


\bibitem[\protect\citeauthoryear{Dai, Zheng, Na, Wang, and Zhang}{Dai et~al\mbox{.}}{2019}]%
        {dai2019eeg}
\bibfield{author}{\bibinfo{person}{Mengxi Dai}, \bibinfo{person}{Dezhi Zheng}, \bibinfo{person}{Rui Na}, \bibinfo{person}{Shuai Wang}, {and} \bibinfo{person}{Shuailei Zhang}.} \bibinfo{year}{2019}\natexlab{}.
\newblock \showarticletitle{EEG classification of motor imagery using a novel deep learning framework}.
\newblock \bibinfo{journal}{\emph{Sensors}} \bibinfo{volume}{19}, \bibinfo{number}{3} (\bibinfo{year}{2019}), \bibinfo{pages}{551}.
\newblock


\bibitem[\protect\citeauthoryear{Duan, Zhu, and Lu}{Duan et~al\mbox{.}}{2013}]%
        {duan2013differential}
\bibfield{author}{\bibinfo{person}{Ruo-Nan Duan}, \bibinfo{person}{Jia-Yi Zhu}, {and} \bibinfo{person}{Bao-Liang Lu}.} \bibinfo{year}{2013}\natexlab{}.
\newblock \showarticletitle{Differential entropy feature for EEG-based emotion classification}. In \bibinfo{booktitle}{\emph{2013 6th International IEEE/EMBS Conference on Neural Engineering (NER)}}. IEEE, \bibinfo{pages}{81--84}.
\newblock


\bibitem[\protect\citeauthoryear{Gajic}{Gajic}{2003}]%
        {gajic2003linear}
\bibfield{author}{\bibinfo{person}{Zoran Gajic}.} \bibinfo{year}{2003}\natexlab{}.
\newblock \bibinfo{booktitle}{\emph{Linear dynamic systems and signals}}.
\newblock \bibinfo{publisher}{Prentice Hall/Pearson Education Upper Saddle River}.
\newblock


\bibitem[\protect\citeauthoryear{Golub, Heath, and Wahba}{Golub et~al\mbox{.}}{1979}]%
        {golub1979generalized}
\bibfield{author}{\bibinfo{person}{Gene~H Golub}, \bibinfo{person}{Michael Heath}, {and} \bibinfo{person}{Grace Wahba}.} \bibinfo{year}{1979}\natexlab{}.
\newblock \showarticletitle{Generalized cross-validation as a method for choosing a good ridge parameter}.
\newblock \bibinfo{journal}{\emph{Technometrics}} \bibinfo{volume}{21}, \bibinfo{number}{2} (\bibinfo{year}{1979}), \bibinfo{pages}{215--223}.
\newblock


\bibitem[\protect\citeauthoryear{Hsu}{Hsu}{2010}]%
        {hsu2010eeg}
\bibfield{author}{\bibinfo{person}{Wei-Yen Hsu}.} \bibinfo{year}{2010}\natexlab{}.
\newblock \showarticletitle{EEG-based motor imagery classification using neuro-fuzzy prediction and wavelet fractal features}.
\newblock \bibinfo{journal}{\emph{Journal of Neuroscience Methods}} \bibinfo{volume}{189}, \bibinfo{number}{2} (\bibinfo{year}{2010}), \bibinfo{pages}{295--302}.
\newblock


\bibitem[\protect\citeauthoryear{Jia, Lin, Wang, Feng, Xie, and Chen}{Jia et~al\mbox{.}}{2021}]%
        {jia2021hetemotionnet}
\bibfield{author}{\bibinfo{person}{Ziyu Jia}, \bibinfo{person}{Youfang Lin}, \bibinfo{person}{Jing Wang}, \bibinfo{person}{Zhiyang Feng}, \bibinfo{person}{Xiangheng Xie}, {and} \bibinfo{person}{Caijie Chen}.} \bibinfo{year}{2021}\natexlab{}.
\newblock \showarticletitle{HetEmotionNet: two-stream heterogeneous graph recurrent neural network for multi-modal emotion recognition}. In \bibinfo{booktitle}{\emph{Proceedings of the 29th ACM International Conference on Multimedia}}. \bibinfo{pages}{1047--1056}.
\newblock


\bibitem[\protect\citeauthoryear{Jia, Lin, Wang, Zhou, Ning, He, and Zhao}{Jia et~al\mbox{.}}{2020}]%
        {jia2020graphsleepnet}
\bibfield{author}{\bibinfo{person}{Ziyu Jia}, \bibinfo{person}{Youfang Lin}, \bibinfo{person}{Jing Wang}, \bibinfo{person}{Ronghao Zhou}, \bibinfo{person}{Xiaojun Ning}, \bibinfo{person}{Yuanlai He}, {and} \bibinfo{person}{Yaoshuai Zhao}.} \bibinfo{year}{2020}\natexlab{}.
\newblock \showarticletitle{GraphSleepNet: Adaptive Spatial-Temporal Graph Convolutional Networks for Sleep Stage Classification.}. In \bibinfo{booktitle}{\emph{IJCAI}}. \bibinfo{pages}{1324--1330}.
\newblock


\bibitem[\protect\citeauthoryear{Koelstra, Muhl, Soleymani, Lee, Yazdani, Ebrahimi, Pun, Nijholt, and Patras}{Koelstra et~al\mbox{.}}{2011}]%
        {koelstra2011deap}
\bibfield{author}{\bibinfo{person}{Sander Koelstra}, \bibinfo{person}{Christian Muhl}, \bibinfo{person}{Mohammad Soleymani}, \bibinfo{person}{Jong-Seok Lee}, \bibinfo{person}{Ashkan Yazdani}, \bibinfo{person}{Touradj Ebrahimi}, \bibinfo{person}{Thierry Pun}, \bibinfo{person}{Anton Nijholt}, {and} \bibinfo{person}{Ioannis Patras}.} \bibinfo{year}{2011}\natexlab{}.
\newblock \showarticletitle{Deap: A database for emotion analysis; using physiological signals}.
\newblock \bibinfo{journal}{\emph{IEEE transactions on affective computing}} \bibinfo{volume}{3}, \bibinfo{number}{1} (\bibinfo{year}{2011}), \bibinfo{pages}{18--31}.
\newblock


\bibitem[\protect\citeauthoryear{Krstajic, Buturovic, Leahy, and Thomas}{Krstajic et~al\mbox{.}}{2014}]%
        {krstajic2014cross}
\bibfield{author}{\bibinfo{person}{Damjan Krstajic}, \bibinfo{person}{Ljubomir~J Buturovic}, \bibinfo{person}{David~E Leahy}, {and} \bibinfo{person}{Simon Thomas}.} \bibinfo{year}{2014}\natexlab{}.
\newblock \showarticletitle{Cross-validation pitfalls when selecting and assessing regression and classification models}.
\newblock \bibinfo{journal}{\emph{Journal of cheminformatics}}  \bibinfo{volume}{6} (\bibinfo{year}{2014}), \bibinfo{pages}{1--15}.
\newblock


\bibitem[\protect\citeauthoryear{Lin, Wang, Wu, Jeng, and Chen}{Lin et~al\mbox{.}}{2007}]%
        {lin2007multilayer}
\bibfield{author}{\bibinfo{person}{Yuan-Pin Lin}, \bibinfo{person}{Chi-Hong Wang}, \bibinfo{person}{Tien-Lin Wu}, \bibinfo{person}{Shyh-Kang Jeng}, {and} \bibinfo{person}{Jyh-Horng Chen}.} \bibinfo{year}{2007}\natexlab{}.
\newblock \showarticletitle{Multilayer perceptron for EEG signal classification during listening to emotional music}. In \bibinfo{booktitle}{\emph{TENCON 2007-2007 IEEE region 10 conference}}. IEEE, \bibinfo{pages}{1--3}.
\newblock


\bibitem[\protect\citeauthoryear{Sanei and Chambers}{Sanei and Chambers}{2013}]%
        {sanei2013eeg}
\bibfield{author}{\bibinfo{person}{Saeid Sanei} {and} \bibinfo{person}{Jonathon~A Chambers}.} \bibinfo{year}{2013}\natexlab{}.
\newblock \bibinfo{booktitle}{\emph{EEG signal processing}}.
\newblock \bibinfo{publisher}{John Wiley \& Sons}.
\newblock


\bibitem[\protect\citeauthoryear{Song, Zheng, Song, and Cui}{Song et~al\mbox{.}}{2018}]%
        {song2018eeg}
\bibfield{author}{\bibinfo{person}{Tengfei Song}, \bibinfo{person}{Wenming Zheng}, \bibinfo{person}{Peng Song}, {and} \bibinfo{person}{Zhen Cui}.} \bibinfo{year}{2018}\natexlab{}.
\newblock \showarticletitle{EEG emotion recognition using dynamical graph convolutional neural networks}.
\newblock \bibinfo{journal}{\emph{IEEE Transactions on Affective Computing}} \bibinfo{volume}{11}, \bibinfo{number}{3} (\bibinfo{year}{2018}), \bibinfo{pages}{532--541}.
\newblock


\bibitem[\protect\citeauthoryear{Srinivasan}{Srinivasan}{2007}]%
        {srinivasan2007cognitive}
\bibfield{author}{\bibinfo{person}{Narayanan Srinivasan}.} \bibinfo{year}{2007}\natexlab{}.
\newblock \showarticletitle{Cognitive neuroscience of creativity: EEG based approaches}.
\newblock \bibinfo{journal}{\emph{Methods}} \bibinfo{volume}{42}, \bibinfo{number}{1} (\bibinfo{year}{2007}), \bibinfo{pages}{109--116}.
\newblock


\bibitem[\protect\citeauthoryear{Vecchiato, Toppi, Astolfi, De~Vico~Fallani, Cincotti, Mattia, Bez, and Babiloni}{Vecchiato et~al\mbox{.}}{2011}]%
        {vecchiato2011spectral}
\bibfield{author}{\bibinfo{person}{Giovanni Vecchiato}, \bibinfo{person}{Jlenia Toppi}, \bibinfo{person}{Laura Astolfi}, \bibinfo{person}{Fabrizio De~Vico~Fallani}, \bibinfo{person}{Febo Cincotti}, \bibinfo{person}{Donatella Mattia}, \bibinfo{person}{Francesco Bez}, {and} \bibinfo{person}{Fabio Babiloni}.} \bibinfo{year}{2011}\natexlab{}.
\newblock \showarticletitle{Spectral EEG frontal asymmetries correlate with the experienced pleasantness of TV commercial advertisements}.
\newblock \bibinfo{journal}{\emph{Medical \& biological engineering \& computing}}  \bibinfo{volume}{49} (\bibinfo{year}{2011}), \bibinfo{pages}{579--583}.
\newblock


\bibitem[\protect\citeauthoryear{Wu, Souza, Zhang, Fifty, Yu, and Weinberger}{Wu et~al\mbox{.}}{2019}]%
        {wu2019simplifying}
\bibfield{author}{\bibinfo{person}{Felix Wu}, \bibinfo{person}{Amauri Souza}, \bibinfo{person}{Tianyi Zhang}, \bibinfo{person}{Christopher Fifty}, \bibinfo{person}{Tao Yu}, {and} \bibinfo{person}{Kilian Weinberger}.} \bibinfo{year}{2019}\natexlab{}.
\newblock \showarticletitle{Simplifying graph convolutional networks}. In \bibinfo{booktitle}{\emph{International conference on machine learning}}. PMLR, \bibinfo{pages}{6861--6871}.
\newblock


\bibitem[\protect\citeauthoryear{Xu, Hu, Leskovec, and Jegelka}{Xu et~al\mbox{.}}{2018}]%
        {xu2018powerful}
\bibfield{author}{\bibinfo{person}{Keyulu Xu}, \bibinfo{person}{Weihua Hu}, \bibinfo{person}{Jure Leskovec}, {and} \bibinfo{person}{Stefanie Jegelka}.} \bibinfo{year}{2018}\natexlab{}.
\newblock \showarticletitle{How powerful are graph neural networks?}
\newblock \bibinfo{journal}{\emph{arXiv preprint arXiv:1810.00826}} (\bibinfo{year}{2018}).
\newblock


\bibitem[\protect\citeauthoryear{Yun, Jeong, Kim, Kang, and Kim}{Yun et~al\mbox{.}}{2019}]%
        {yun2019graph}
\bibfield{author}{\bibinfo{person}{Seongjun Yun}, \bibinfo{person}{Minbyul Jeong}, \bibinfo{person}{Raehyun Kim}, \bibinfo{person}{Jaewoo Kang}, {and} \bibinfo{person}{Hyunwoo~J Kim}.} \bibinfo{year}{2019}\natexlab{}.
\newblock \showarticletitle{Graph transformer networks}.
\newblock \bibinfo{journal}{\emph{Advances in neural information processing systems}}  \bibinfo{volume}{32} (\bibinfo{year}{2019}).
\newblock


\bibitem[\protect\citeauthoryear{Zhang, Yu, Liu, Zhao, Zhang, and Zheng}{Zhang et~al\mbox{.}}{2021a}]%
        {zhang2021sparsedgcnn}
\bibfield{author}{\bibinfo{person}{Guanhua Zhang}, \bibinfo{person}{Minjing Yu}, \bibinfo{person}{Yong-Jin Liu}, \bibinfo{person}{Guozhen Zhao}, \bibinfo{person}{Dan Zhang}, {and} \bibinfo{person}{Wenming Zheng}.} \bibinfo{year}{2021}\natexlab{a}.
\newblock \showarticletitle{SparseDGCNN: Recognizing emotion from multichannel EEG signals}.
\newblock \bibinfo{journal}{\emph{IEEE Transactions on Affective Computing}} (\bibinfo{year}{2021}).
\newblock


\bibitem[\protect\citeauthoryear{Zhang, Zhao, Wei, Mantini, Li, and Liu}{Zhang et~al\mbox{.}}{2021b}]%
        {zhang2021eegdenoisenet}
\bibfield{author}{\bibinfo{person}{Haoming Zhang}, \bibinfo{person}{Mingqi Zhao}, \bibinfo{person}{Chen Wei}, \bibinfo{person}{Dante Mantini}, \bibinfo{person}{Zherui Li}, {and} \bibinfo{person}{Quanying Liu}.} \bibinfo{year}{2021}\natexlab{b}.
\newblock \showarticletitle{Eegdenoisenet: A benchmark dataset for deep learning solutions of eeg denoising}.
\newblock \bibinfo{journal}{\emph{Journal of Neural Engineering}} \bibinfo{volume}{18}, \bibinfo{number}{5} (\bibinfo{year}{2021}), \bibinfo{pages}{056057}.
\newblock


\bibitem[\protect\citeauthoryear{Zhang, Pan, Shen, ud~Din, Li, Lu, Wu, and Hu}{Zhang et~al\mbox{.}}{2020}]%
        {zhang2020fusing}
\bibfield{author}{\bibinfo{person}{Xiaowei Zhang}, \bibinfo{person}{Jing Pan}, \bibinfo{person}{Jian Shen}, \bibinfo{person}{Zia ud Din}, \bibinfo{person}{Junlei Li}, \bibinfo{person}{Dawei Lu}, \bibinfo{person}{Manxi Wu}, {and} \bibinfo{person}{Bin Hu}.} \bibinfo{year}{2020}\natexlab{}.
\newblock \showarticletitle{Fusing of electroencephalogram and eye movement with group sparse canonical correlation analysis for anxiety detection}.
\newblock \bibinfo{journal}{\emph{IEEE Transactions on Affective Computing}} \bibinfo{volume}{13}, \bibinfo{number}{2} (\bibinfo{year}{2020}), \bibinfo{pages}{958--971}.
\newblock


\bibitem[\protect\citeauthoryear{Zhong, Wang, and Miao}{Zhong et~al\mbox{.}}{2020}]%
        {zhong2020eeg}
\bibfield{author}{\bibinfo{person}{Peixiang Zhong}, \bibinfo{person}{Di Wang}, {and} \bibinfo{person}{Chunyan Miao}.} \bibinfo{year}{2020}\natexlab{}.
\newblock \showarticletitle{EEG-based emotion recognition using regularized graph neural networks}.
\newblock \bibinfo{journal}{\emph{IEEE Transactions on Affective Computing}} \bibinfo{volume}{13}, \bibinfo{number}{3} (\bibinfo{year}{2020}), \bibinfo{pages}{1290--1301}.
\newblock


\end{thebibliography}

\end{document}